\documentclass[11pt,a4paper]{article}
\usepackage[hyperref]{acl2021}
\usepackage{times}
\usepackage{latexsym}

\newcommand{\T}{{\scriptscriptstyle \top}}
\usepackage{algorithm}
\usepackage{url}
\usepackage{hyperref}
\usepackage{algpseudocode}
\usepackage{makecell}
\usepackage{graphicx}
\usepackage{tabularx}
\usepackage{pgf}
\usepackage{mathtools}
\usepackage{array}
\usepackage{amsmath, amssymb, amsfonts}
\usepackage{array, multirow, booktabs}
\graphicspath{ {./figures/} }
\usepackage{microtype}

\usepackage{microtype}

\aclfinalcopy 

\title{Denoising Word Embeddings by Averaging in a Shared Space}

\author{Avi Caciularu$^1$~~~~~Ido Dagan$^1$~~~~~Jacob Goldberger$^2$\\
    $^1$Computer Science Department, Bar-Ilan University\\
    $^2$Faculty of Engineering, Bar-Ilan University\\
    {\tt\small avi.c33@gmail.com, dagan@cs.biu.ac.il, jacob.goldberger@biu.ac.il} \\
}

\date{}

\begin{document}
\maketitle
\begin{abstract}
We introduce a new approach for smoothing and improving the quality of word embeddings. We consider a method of fusing word embeddings that were trained on the same corpus but with different initializations. We project all the models to a shared vector space using an efficient implementation of the Generalized Procrustes Analysis (GPA) procedure, previously used in multilingual word translation. Our word representation demonstrates consistent improvements over the raw models as well as their simplistic average, on a range of  tasks. As the new representations are more stable and reliable, there is a noticeable improvement in rare word evaluations.
\end{abstract}

\section{Introduction}

Continuous (non-contextualized) word embeddings have been introduced several years ago as
a standard building block for NLP
tasks. These models provide efficient ways to learn word representations in a fully self-supervised manner from text corpora, solely based on word co-occurrence statistics.
A wide variety of methods now exist for generating word embeddings, with prominent methods including word2vec \cite{NIPS2013_5021}, GloVe \cite{pennington2014glove}, and FastText \cite{bojanowski2017enriching}.
Recently, contextualized embeddings \cite{peters-etal-2018-deep,devlin-etal-2019-bert}, replaced the use of non-contextualized embeddings in many settings. Yet, the latter remain the standard choice for typical lexical-semantic tasks, e.g., semantic similarity~\cite{hill2015simlex}, word analogy~\cite{jurgens2012semeval}, relation classification~\cite{barkan-etal-2020-within}, and paraphrase identification~\cite{meged-etal-2020-paraphrasing}. These tasks consider the generic meanings of lexical items, given out of context, hence the use of non-contextualized embeddings is appropriate. Notably, FastText was shown to yield state-of-the-art results in most of these tasks \cite{bojanowski2017enriching}.

While word embedding methods proved to be powerful, they suffer from a certain level of noise, introduced by quite a few randomized steps in the embedding generation process, including embedding initialization, negative sampling, subsampling and mini-batch ordering. Consequently, different runs would yield different embedding geometries, of varying quality. This random noise might harm most severely  the representation of rare words, for which the actual data signal is rather weak~\cite{barkan-etal-2020-bayesian}.

In this paper, we propose denoising word embedding models through generating multiple model versions, each created with different random seeds. 
Then, the resulting representations for each word should be fused effectively, in order to obtain a model with a reduced level of noise.
Note, however, that simple averaging of the original word vectors is problematic, since each training session of the algorithm produces embeddings in a different space. In fact, the objective scores of both word2vec, Glove and FastText are invariant to multiplying all the word embeddings by an orthogonal matrix, hence, the algorithm output involves an arbitrary rotation of the embedding space. 

For addressing this issue, we were inspired by recent approaches originally proposed for aligning multi-lingual embeddings \cite{Chen2018,Yova_2018,Alaux2019,Pratik2019, Hagai2019}.
To obtain such alignments, these methods simultaneously project the original language-specific embeddings into a shared space, while enforcing (or at least encouraging) transitive orthogonal transformations. 
In our (monolingual) setting, we propose a related technique to project the different embedding versions into a shared space, while optimizing the projection towards obtaining an improved fused representation. We show that this results in improved performance on a range of lexical-semantic tasks, with notable improvements for rare words, as well as on several sentence-level downstream tasks.

\section{Word Averaging in a Shared Space}
Assume we are given an ensemble of $k$ pre-trained word embedding sets, of the same word vocabulary of size $n$ and the same dimensionality $d$. In our setting, these sets are obtained by training the same embedding model using different random parameter initializations. Our goal is to fuse the $k$ embedding sets into a single ``average'' embedding that is hopefully more robust and would yield better performance on various tasks. Since each embedding set has its own space, we project the $k$ embedding spaces into a shared space, in which we induce averaged embeddings based on a mean squared error minimization objective.

Let $x_{i,t}\in \mathbb{R}^d$ be the dense representation of the $t$-th word in the $i$-th  embedding set.
We model the mapping from the $i$-th set to the shared space by an orthogonal matrix denoted by $T_i$.
Denote the sought shared space representation of the $t$-th word by $y_t\in \mathbb{R}^d$. 
Our goal is to find a set of transformations $T=\left\{T_1,...,T_k\right\}$ and target word embeddings $y=\left\{y_1,...,y_n\right\}$ in the shared space that minimize the following mean-squared error:
\begin{equation}
S(T,y)  = \sum_{i=1}^k \sum_{t=1}^n \left\|T_i x_{i,t}-y_t\right\|^2.
\label{score}
\end{equation}
For this objective, it is easy to show that for a set of transformations $T_1,...,T_k$,
the optimal shared space representation is: $$y_t = \frac{1}{k} \sum_{i=1}^k T_i x_{i,t}.$$ Hence, solving the optimization problem pertains to finding the $k$ optimal transformations.

In the case where $k=2$, the optimal $T$ can be obtained in a closed form using the Procrustes Analysis (PA) procedure \cite{procrustes1966}, which has been employed in recent bilingual word translation methods \cite{xing2015normalized, artetxe2016learning,hamilton2016diachronic,artetxe2017acl,artetxe2017unsupervised,conneau2017word,artetxe2018generalizing,artetxe2018robust, ruder2018discriminative}. In our setting, to obtain an improved embedding, we wish to average more than two embedding sets.


However, if $k>2$ there is no closed form solution to (\ref{score}) and thus, we need to find a solution using an iterative optimization process. To that end, we follow several works that suggested employing the General Procrustes Analysis (GPA) procedure, which is an extension of PA to multi-set alignment \cite{Gower1975,Yova_2018}. 
Generally, the GPA consists of an alternate minimization procedure where we iterate between finding the orthogonal transformations and computing the shared space.
The optimal transformation from each embedding space to the shared space is found by minimizing the following score,
\begin{equation*}
S(T_i)  = \sum_{t=1}^n \left\|T_i x_{i,t}-y_t\right\|^2,  \hspace{0.5cm} i=1,...,k.
\label{score_ti}
\end{equation*}

The minimum of $S(T_i)$ can then be found by the closed form PA procedure.  The updated transformation is $T_i = U_iV_i^{\T}$, where $ U_i\Sigma_i V^{\T}_i $  is the singular value decomposition (SVD) of the $ d \times d $ matrix  $ \sum_{t=1}^n y_{t}^{}x_{i,t}^{\T}$. 
At each step in the iterative GPA algorithm, the score (\ref{score}) is monotonically decreased until it converges to a local minimum point. 
\begin{algorithm}[th]
\begin{algorithmic}[1]
\State \textbf{Input:} Ensemble of $k$ word embedding sets.
\State \textbf{Task:}  Find the optimal average embedding.
\State {\bf Preprocessing:} 
\State Compute the cross-correlation matrices: \\
  $ C_{ij}^{}= C_{ji}^{\T} =  \sum_{t=1}^n x_{j,t}^{}x_{i,t}^{\T}  \hspace{0.5cm} 1\le i  < j \le k$
\State{\textbf{Initialization:}} $ T_1=  \cdots = T_{k-1}=0,$ \hspace{-0.1cm} $T_k=I$ 
\While{not converged}
\For{$ i =1,...,k$}
\State $ U\Sigma V^{\T} = \text{SVD}\left(\sum_{j\ne i}  T_j C_{ij}\right)$
\State $T_i \leftarrow UV^{\T}$
\EndFor
\EndWhile
\State \textbf{Compute the average embedding:}
\State $y_t \leftarrow \frac{1}{k} \sum_{i=1}^k T_i x_{i,t} \hspace{1cm} t=1,...,n$
\end{algorithmic}
\caption{Shared Space Embedding Averaging} 
\label{algo1}
\end{algorithm}

For large vocabularies, GPA is not efficient, because, in each iteration, when computing the SVD we need to sum over all the vocabulary words. To circumvent this computational cost, we adopt the optimization procedure from \citet{Hagai2019}, which we apply within each iteration. Instead of summing over the whole vocabulary, the following extension is proposed. Let $C_{ij} = \sum_t x_{j,t} x_{i,t}^{\T}$ be the cross-correlation matrix for a pair $(i,j)$ of two original embedding spaces, which can be computed once, for all pairs of spaces, in a pre-processing step. Given the matrices $C_{ij}$ the 
computational complexity of the iterative averaging algorithm is independent of the vocabulary size, allowing us to compute efficiently the SVD. The resulting algorithm termed Shared Space Embedding Averaging (SSEA) is presented in Algorithm~\ref{algo1}.\footnote{The algorithm demonstration code is available at \href{https://github.com/aviclu/SSEA}{github.com/aviclu/SSEA}. In practice, we utilized an efficient PyTorch implementation based on~\citet{Hagai2019}.}

\section{Experimental Setup and Results}
This section presents our evaluation protocol, datasets, data preparation, hyperparameter configuration and results. 

\subsection{Implementation Details and Data}
We trained word2vec \cite{NIPS2013_5021}, FastText \cite{bojanowski2017enriching} and GloVe \cite{pennington2014glove} embeddings. 
For word2vec we used the skip-gram model with negative sampling, which was shown advantageous 
on the evaluated tasks 
\cite{levy2015improving}. We trained each of the models on the November 2019 dump of Wikipedia articles\footnote{\href{https://dumps.wikimedia.org/enwiki/latest/}{dumps.wikimedia.org/enwiki/latest/}} for $k=30$ times, with different random seeds, and used the default reported hyperparameters; we set the embedding dimension to $d=200$, and considered each word within the maximal window $c_{max} = 5$, subsampling\footnote{To speed up the training.} threshold of $\rho=10^{-5}$ and used $5$ negative examples for every positive example. In order to keep a large amount of rare words in the corpus, no preprocessing was applied on the data, yielding a vocabulary size of $1.5\cdot 10^6$. We then applied the SSEA algorithm to the embedding sets to obtain the average embedding. The original embedding sets and averaged embeddings were centered around the $0$ vector and normalized to unit vectors.

\begin{table}[t!]
\centering
\footnotesize
\begin{tabular}{@{}lcc}
\toprule
              & original & denoised  \\\midrule
 word2vec      & $0.40 \pm 0.005$   & $0.059 \pm 0.003$   \\
GloVe      & $0.38 \pm 0.006$   & $0.058 \pm 0.003$ \\
FastText & $0.35 \pm 0.003$   & $0.054 \pm 0.001$    \\ \bottomrule
\end{tabular}
\caption{Average MSE scores of the embedding models with and without applying the SSEA algorithm.}
\label{tbl:mse}
\vskip -0.1in
\end{table}

  \begin{figure}[t!]
    \centering
    \includegraphics[width=0.5\textwidth]{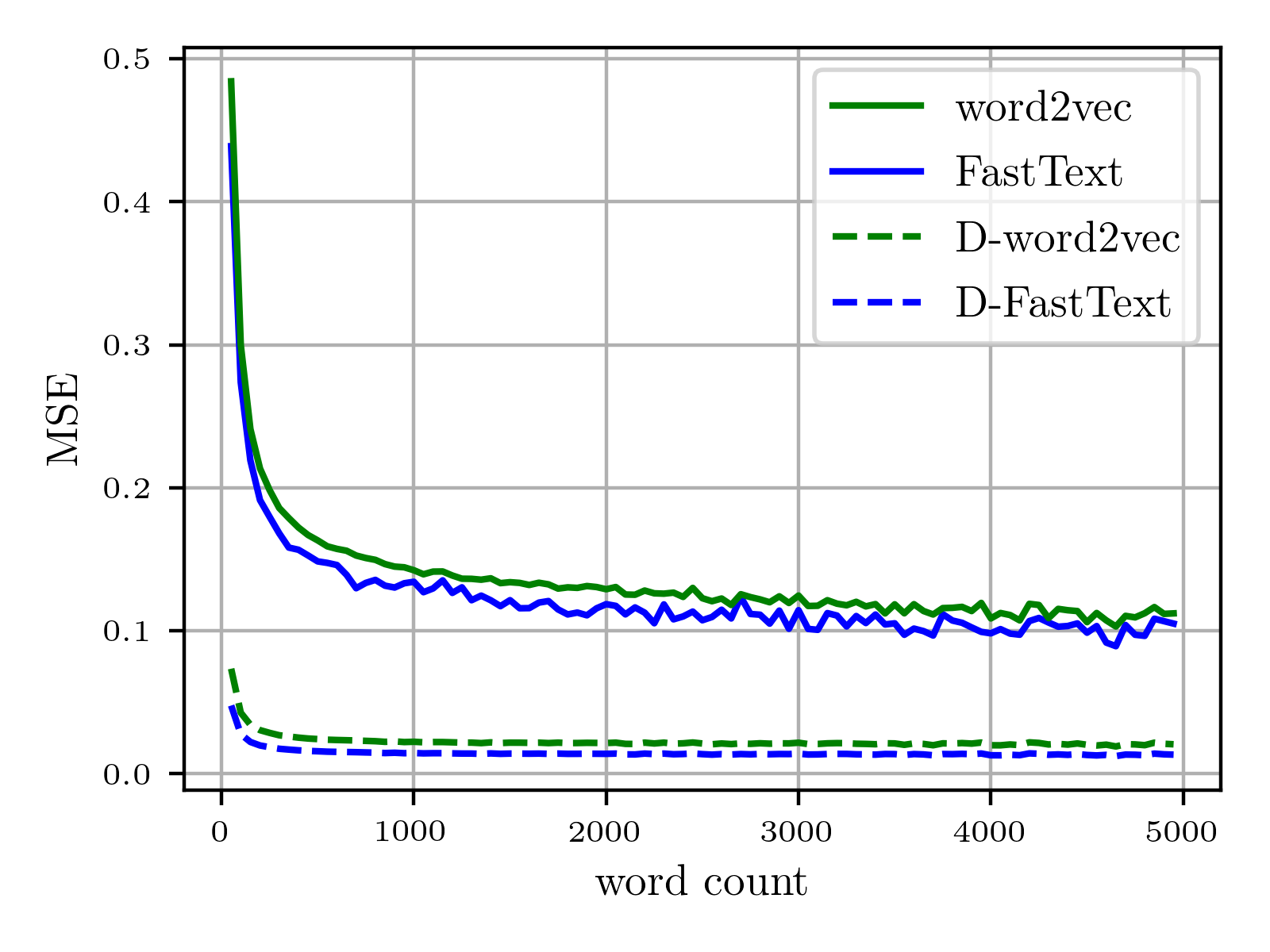}
    \vskip -0.2in
    \caption{Average MSE for word embeddings vs their corpus occurrence count (binned with resolution of $50$).}
    \label{fig:counts}
\vskip -0.1in
\end{figure}

\begin{table*}[t!]
    \centering 
    \footnotesize
    \begin{tabularx}{\textwidth}{p{17mm} p{11.9mm}p{11.1mm}p{11mm}p{11.1mm}p{11.5mm}p{11.1mm}p{18.1mm}p{11.1mm}p{11.1mm}}
 
 \toprule
Method &  SimLex999 &    MEN &  WS353 &     AP &  Google &    MSR &  SemEval2012(2) &  BLESS &  RW \\
\midrule
${\text{word2vec}}$ &      {33.7} &   {72.4} &  {60.7} &{62.2} &{69.5} &  {51.3} &          {19.2} &  {79} &  {42}  \\
${\text{A-word2vec}}$       &      33.1 &   72.3 &  \bf 60.8 &  61.9 &  69.2 & 51.2 & 19 &  78.1 &  41.7\\
${\text{D-word2vec}}$  &\bf 33.9 &  \bf 73.2 & \bf 60.8 &  \bf 63.1 &   \bf 70.3 &  \bf 51.9 &   \bf 20 &  \bf 79.6 & \bf  43.4 \\
\midrule
${\text{GloVe}}$       &     {34.4} & {73.4}  &  {62.3} &  {63.3} &   {75.1} &  {54.5} & {19.7} &  {79.2} &{47.2}  \\
${\text{A-GloVe}}$         & 34.2 &  73.1 &  61.9 & 62.8 &  74.7 & 54.2 & 19.6 & 79 &  47.1 \\
${\text{D-GloVe}}$         &      \bf 34.8 &  \bf 75.1 &  \bf 62.7 &  \bf 64.3 &  \bf 75.9 & \bf 55.2 & \bf 20.1 & \bf 79.9 & \bf  48.5 \\
\midrule
${\text{FastText}}$       &     {41.2}  &  {78.6} &{70.7} &  {72.2} &   {75.7} &  {63.4} & {19.8} &  {81.5} &  {47.1} \\
${\text{A-FastText}}$    &   41 &  78.1 &  69.7 & 72.1  &   74.1 &  62.8 &  19.4 &  80.8 &  46.6 \\
${\text{D-FastText}}$         &     \bf 42.2 &  \bf 79.3 &  \bf 71.8 &  \bf 72.9 &   \bf 77.4 &  \bf 63.8 &          \bf 20.2 &  \bf 82.7 & \bf 50.3  \\
\bottomrule
    \end{tabularx}
    \caption{Results for lexical-semantic benchmarks. Best performance is bolded.}
    \label{tab:results-table}
\vspace{-1.em}
\end{table*}
\subsection{Improved Embedding Stability}
We next analyze how our method improves embedding quality and consistency, notably for rare words. To that end, for any two embedding sets, $u$ and $v$, we can find the optimal mapping $Q$ between them using the PA algorithm and compute its mean square error (MSE), $\frac{1}{n}\sum_{t=1}\|Qu_t-v_t\|^2$. We define the stability of an embedding algorithm by the average MSE (over $10$ random pairs of samples) between two instances of it. This score measures the  similarity between the geometries of random instances generated by a particular embedding method , and thus reflects the consistency and stability of that method. The scores of the different models are depicted in Table~\ref{tbl:mse}. As observed, after applying SSEA the Average MSE drops by an order of magnitude, indicating much better stability of the obtained embeddings.

We can perform a similar analysis for each word separately. A consistent embedding of the $t$-th word in both sets $u$ and $v$ should result in a small mapping discrepancy $\|Qu_t-v_t\|^2$. Figure \ref{fig:counts} depicts MSE for the models and their computed SSEA, as a function of  the word's frequency in the corpus. The denoised version of the models is marked with a ‘D-’ prefix. For clarity of presentation, we did not include the results for GloVe (which are similar to word2vec). 
As expected, embedding stability always increases (MSE decreases) with word frequency. SSEA is notably more stable across the frequency range, with the error minimized early on and reduced most drastically for low frequencies.


\subsection{Comparison of methods}
We next compare our denoised model, denoted with a ‘D-’ prefix, with the original embedding models. 
As an additional baseline, we considered also the na\"ive averaged embedding model, denoted with a ‘A-’ prefix, where for every word we computed the simplistic mean embedding across all original spaces. 
Note that we did not compare other proposed embeddings or meta-embedding learning methods, but rather restricted our analysis to empirically verifying our embedding aggregation method and validating the assumptions behind the empirical analysis we performed.  

\subsection{Evaluations on Lexical Semantic Tasks}
We evaluated the performance of our method over lexical-semantic tasks, including 
word similarity, analogy solving, and concept categorization: 
\textbf{SimLex999} \cite{hill2015simlex}, \textbf{MEN} \cite{bruni2014multimodal}, \textbf{WS353} \cite{finkelstein2002placing}, \textbf{AP} \cite{almuhareb2004attribute}, \textbf{Google} \cite{mikolov2013distributed}, \textbf{MSR} \cite{mikolov2013linguistic}, \textbf{SemEval-2012} \cite{jurgens2012semeval}, \textbf{BLESS} \cite{baroni2011we} and \textbf{RW} \cite{luong2013better}, (focusing on
rare words). For the analogy task, we reported the accuracy. For the remaining tasks, we computed Spearman’s correlation between the cosine similarity of the embeddings and the human judgments.

{\bf Results}
The results of the lexical-semantic tasks are depicted in Table \ref{tab:results-table}, averaged over 30 runs for each method.
Our method obtained better performance than the other methods, substantially for FastText embeddings. As shown, the na\"ive averaging performed poorly, which highlights the fact that simply averaging different embedding spaces does not improve word representation quality. The most notable performance gain was in the rare-words task, in line  with the analysis in Fig.~\ref{fig:counts}, suggesting that on rare words the raw embedding vectors fit the data less accurately. 

\subsection{Evaluations On Downstream Tasks}
\label{downstream}
For completeness, we next show the relative advantage of our denoising method also when applied to several sentence-level downstream benchmarks. While contextualized embeddings dominate a wide range of sentence- and document- level NLP tasks \cite{peters-etal-2018-deep,devlin-etal-2019-bert,caciularu2021cross}, we assessed the relative advantage of our denoising method when utilizing (non-contextualized) word embeddings in sentence- an document- level settings. We applied the exact procedure proposed in \citet{li-etal-2017-investigating} and \citet{rogers-etal-2018-whats}, as an effective benchmark for the quality of static embedding models. We first used sequence labeling tasks. The morphological and syntactic performance was evaluated using part-of-speech tagging, \textbf{POS}, and chunking, \textbf{CHK}. Both named entity recognition, \textbf{NER}, and multi-way classification of semantic relation classes, \textbf{RE}, tasks were used for evaluating semantic information at the word level. For the above POS, NER and CHK sequence labeling tasks, we used the CoNLL 2003 dataset \cite{tjongkimsang2003conll} and for the RE task, we used the SemEval 2010 task 8 dataset \cite{hendrickx-etal-2010-semeval}. The neural network models employed for these downstream tasks are fully described in \cite{rogers-etal-2018-whats}. Next, we evaluated the following semantic level tasks: document-level polarity classification, \textbf{PC}, using the Stanford IMDB movie review dataset \cite{maas-etal-2011-learning}, sentence level sentiment polarity classification, \textbf{SEN}, using the MR dataset of short movie reviews \cite{pang-lee-2005-seeing}, and classification of subjectivity and objectivity task, \textbf{SUB}, that uses the Rotten Tomatoes user review snippets against official movie plot summaries \cite{pang-lee-2004-sentimental}. Similarly to the performance results in Table~\ref{tab:results-table}, the current results show that the suggested denoised embeddings obtained better overall performance than the other methods, substantially for FastText embeddings.

\begin{table}[t]
    \centering
    \footnotesize
    \begin{tabular}{p{13mm}p{4.1mm}p{4.1mm}p{4.1mm}p{4.1mm}p{4.1mm}p{4.1mm}p{4.1mm}}
 
 \toprule
Method & POS & CHK & NER & RE & PC & SEN & SUB \\
\midrule
${\text{word2vec}}$ &{81.5} & {80.1} & {93.3} & {71.4} & {89.2} &{73.9} & {76.4} \\
${\text{A-word2vec}}$       & 78  & 77.5  & 90.9  & 67.4  & 86.4 & 64.3 & 75.6 \\
${\text{D-word2vec}}$ & \textbf{81.6} & \textbf{80.2} & \textbf{93.6} & \textbf{73.1}& \textbf{89.7} & \textbf{74} & \textbf{77.4} \\
\midrule
${\text{GloVe}}$        & {77.5} & {70.4} & {85.2} & {66.7} & {80.2} &{70.2} &{72.7}\\
${\text{A-GloVe}}$  &77.1  & 70.2 & 84.9 & 62.3 &77.7 &62.2 & 71.8\\
${\text{D-GloVe}}$  & \textbf{77.8} & \textbf{71.1} & \textbf{86.6} & \textbf{68.2} & \textbf{80.8} & \textbf{71.3} & \textbf{73.9} \\
\midrule
${\text{FastText}}$ & {80.6} & {79.1} &  {92.2}& {74} &88.9 & {74.9}& {73.9}\\
${\text{A-FastText}}$ & 78.4  & 78.8 & 90.2  & 73.6  & 89 & 74.1 & 73.3 \\
${\text{D-FastText}}$ & \textbf{82.4} & \textbf{81.2} & \textbf{94.9} & \textbf{75.2} & \textbf{90.5} & \textbf{77.3} & \textbf{76.7} \\
\bottomrule

    \end{tabular}
    \caption{Results for downstream task. Best performance is bolded.}
    \label{tab:results-downstream-table}
\vspace{-2.mm}
\end{table}


\section{Related Work}

A similar situation of aligning different word embeddings into a shared space occurs in multi-lingual word translation tasks which are 
based on distinct monolingual word embeddings. Word translation is performed by transforming each language word embeddings into a shared space by an orthogonal matrix, for creating a ``universal language'', which is useful for the word translation process. Our setting may be considered by viewing each embedding set as a different language, where our goal is to find the shared space where embedding averaging is meaningful.

The main challenge in multilingual word translation is to obtain a reliable multi-way word correspondence in either a supervised or unsupervised manner. One problem is that standard dictionaries contain multiple senses for words, which is problematic for bilingual translation, and further amplified in a multilingual setting. In our case of embedding averaging, the mapping problem vanishes since we are addressing a single language and the word  correspondences hold trivially among different embeddings of the same word. Thus, in our setting, there are no problems of wrong word correspondences, neither the issue of having different word translations due to multiple word senses. Studies have shown that for the multi-lingual translation  problem, enforcing the transformation to be strictly orthogonal is too restrictive and performance can be improved by using the  orthogonalization as a regularization \cite{Chen2018} that yields matrices that are close to be orthogonal. In our much simpler setting of a single language, with a trivial identity word correspondence, enforcing the  orthogonalization constraint is reasonable.

Another related problem is \textit{meta-embedding} \cite{YinS16}, which aims to fuse information from different embedding models. Various methods have been proposed for embedding fusion, such as concatenation, simple averaging, weighted averaging \cite{Coates2018,kiela2018dynamic} and autoencoding \cite{BollegalaB18}. Some of these methods (concatenation and autoencoding) are not scalable when the goal is to fuse many sets, while others (simple averaging) yield inferior results, as described in the above works. Note that our method is not intended to be a competitor of meta-embedding, but rather a complementary method.

An additional related work is the recent method from~\cite{muromagi-etal-2017-linear}. Similarly to our work, they proposed a method based on the Procrustes Analysis procedure for aligning and averaging sets of word embedding models. However, the mapping algorithm they used is much more computationally demanding, as it requires to go over all the dictionary words in every iteration. Instead, we propose an efficient optimization algorithm, which requires just one such computation during each iteration, and is theoretically guaranteed to converge to a local minimum point. While their work focuses on improving over the Estonian language, we suggest evaluating this approach on English data and on a range of different downstream tasks. We show that our method significantly improves upon rare words, which is beneficial for small sized / domain-specific corpora.

\section{Conclusions}
We presented a novel technique for creating better word representations by training an embedding model several times, from which we derive an averaged representation. The resulting word representations proved to be more stable and reliable than the raw embeddings. Our method exhibits performance gains in lexical-semantic tasks, notably over rare words, confirming our analytical assumptions. This suggests that our method may be particularly useful for training embedding models in low-resource settings. Appealing future research may extend our approach to improving sentence-level representations, by fusing several contextualized embedding models.

\section*{Acknowledgments}
The authors would like to thank the anonymous reviewers for their comments and suggestions. 
The work described herein was supported in part by grants from Intel Labs, Facebook, the Israel Science Foundation grant 1951/17 and the German Research Foundation through the German-Israeli Project Cooperation (DIP, grant DA 1600/1-1).

\bibliography{anthology,acl2020}
\bibliographystyle{acl_natbib}

\end{document}